\documentclass[11pt,a4paper]{article}
\usepackage[hyperref]{acl2021}
\usepackage{times}
\usepackage{latexsym}

\usepackage{microtype}

\aclfinalcopy 
\def\aclpaperid{878} 

\usepackage{algpseudocode} 
\usepackage{algorithmicx,algorithm} 
\usepackage{bm} 
\usepackage{multirow} 
\usepackage{booktabs} 
\usepackage{graphicx} 
\usepackage{amssymb} 
\usepackage[utf8]{inputenc}
\usepackage[english]{babel}
\usepackage{color}

\title{Towards Quantifiable Dialogue Coherence Evaluation}

\author{First Author \\
  Affiliation / Address line 1 \\
  Affiliation / Address line 2 \\
  Affiliation / Address line 3 \\
  \texttt{email@domain} \\\And
  Second Author \\
  Affiliation / Address line 1 \\
  Affiliation / Address line 2 \\
  Affiliation / Address line 3 \\
  \texttt{email@domain} \\}

\author{Zheng Ye\textsuperscript{\rm 1}, Liucun Lu\textsuperscript{\rm 1}, Lishan Huang\textsuperscript{\rm 2}, Liang Lin\textsuperscript{\rm 2,3}, Xiaodan Liang\textsuperscript{\rm 1}\thanks{\ \ Corresponding Author.}\\
\textsuperscript{\rm 1}Shenzhen Campus of Sun Yat-sen University, 
\textsuperscript{\rm 2}Sun Yat-Sen University, 
\textsuperscript{\rm 3}Dark Matter AI Inc.\\
\{yezh7,lulc,huanglsh6\}@mail2.sysu.edu.cn, \\ linliang@ieee.org, xdliang328@gmail.com
}

\date{}

\begin{document}
\maketitle
\begin{abstract}
Automatic dialogue coherence evaluation has attracted increasing attention and is crucial for developing promising dialogue systems.
However, existing metrics have two major limitations: 
(a) they are mostly trained in a simplified two-level setting (coherent vs.\ incoherent), while humans give Likert-type multi-level coherence scores, dubbed as ``quantifiable"; (b) their predicted coherence scores cannot align with the actual human rating standards due to the absence of human guidance during training. To address these limitations, we propose \textbf{Quanti}fiable \textbf{D}ialogue \textbf{C}oherence \textbf{E}valuation (QuantiDCE), a novel framework aiming to train a quantifiable dialogue coherence metric that can reflect the actual human rating standards. Specifically, QuantiDCE includes two training stages, Multi-Level Ranking (MLR) pre-training and Knowledge Distillation (KD) fine-tuning. During MLR pre-training, a new MLR loss is proposed for enabling the model to learn the coarse judgement of coherence degrees. Then, during KD fine-tuning, the pretrained model is further finetuned to learn the actual human rating standards with only very few human-annotated data. To advocate the generalizability even with limited fine-tuning data, a novel KD regularization is introduced to retain the knowledge learned at the pre-training stage. Experimental results show that the model trained by QuantiDCE presents stronger correlations with human judgements than the other state-of-the-art metrics. \footnote{The code and trained checkpoints are available at \url{https://github.com/James-Yip/QuantiDCE}.}

\end{abstract}


\begin{figure}[t] 
	\centerline{\includegraphics[width=0.8\linewidth]{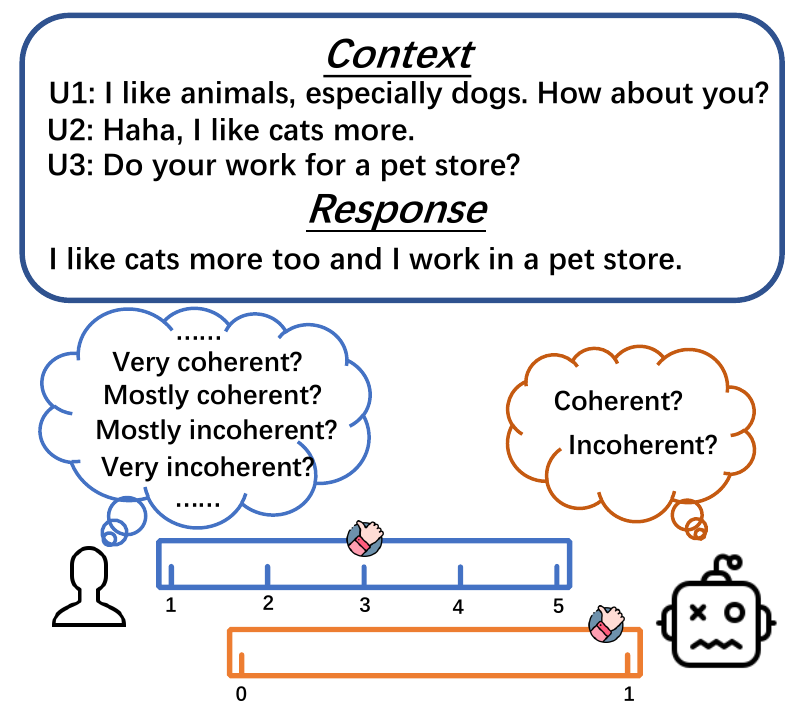}}
	\vspace{-3mm}
    \caption{Likert-type multi-level human rating vs. two-level automatic evaluation. Human rating always considers multiple coherence degrees, while most of the existing automatic metrics only learn to distinguish the coherence dialogues from the incoherent ones and give relatively extreme coherence scores.}
	\label{fig:exam}
	\vspace{-3mm}
\end{figure}

\section{Introduction}
\label{sec:introduction}
Dialogue coherence, which requires a response to be fluent, consistent and context-related, is an essential property for developing promising dialogue systems~\citep{Cervone2018}.
However, it is still challenging to evaluate the coherence of a response generated by a dialogue system.
Although human evaluation is always considered as the most accurate way to evaluate the coherence, it is expensive and high-latency, which cannot meet the evaluation demand of the frequent development of dialogue systems. Therefore, automatic evaluation metrics are developed to serve as human proxies that can rapidly compute the dialogue coherence and return relatively accurate results.

The current widely used metrics measure the lexical word-overlap between generated responses and reference responses, such as BLEU~\citep{bleu} and ROUGE~\citep{rouge}. However, they have been demonstrated to be biased and correlate poorly with human judgements since no semantic information is considered~\citep{liu-etal-2016-evaluate,novikova-etal-2017-need}. 
To overcome this issue, researchers turned to develop learnable metrics based on neural networks that incorporate the semantic information, such as RUBER~\citep{ruber}, BERT-RUBER~\citep{bert-ruber} and GRADE~\citep{grade}. However, these metrics deviate from the actual human rating due to two limitations.
First, they simplify the coherence evaluation task in a two-level setting, i.e., coherent or incoherent, by maximizing the differences between the positive coherent dialogues and the negative incoherent ones obtained by some negative sampling strategies. In contrast, humans usually adopt Likert scaling and give coherence scores from multiple levels like 1 to 5, as shown in Figure~\ref{fig:exam}.
Second, to avoid relying on large-scale human-annotated data, they are mostly trained in a purely unsupervised manner and cannot align with the human rating due to the absence of introducing the actual human rating standards during training.


To address the above limitations, we propose a novel dialogue coherence metric training framework, named as \textbf{Quanti}fiable \textbf{D}ialogue \textbf{C}oherence \textbf{E}valuation (QuantiDCE).
This framework consists of two training stages: Multi-Level Ranking (MLR) pre-training and Knowledge Distillation (KD) fine-tuning. At the MLR pre-training stage, a new multi-level ranking (MLR) loss is proposed for learning the coarse judgement of coherence degrees. Specifically, the MLR loss separates the context-response pairs with different coherence levels and compacts the pairs within the same level in one-dimensional score space. As a result, the pretrained model is able to distinguish different coherence-level dialogue responses for a given context and predicts more accurate coherence scores. At the KD fine-tuning stage, the pretrained model is further finetuned to learn the actual human rating standards with only very few human-annotated coherence scores. To mitigate overfitting into the scarce annotated data during fine-tuning,
a novel knowledge distillation regularization loss is introduced to retain the knowledge learned at the pre-training stage, where the pretrained model (teacher) provides the soft targets for the model during fine-tuning (student). Experimental results show that the metric trained by our QuantiDCE obviously outperforms the other state-of-the-art metrics in terms of the Pearson, Spearman and Kendall correlations with human judgements by around 5\% points on average. To summarize our contributions: 

1) We propose QuantiDCE, a novel quantifiable training framework for dialogue coherence evaluation, which aims to align the automatic scores with the actual human rating standards via MLR pre-training and KD fine-tuning.
To the best of our knowledge, it is the first attempt to consider the quantifiable problem for dialogue coherence evaluation.


2) Extensive experiments demonstrate the effectiveness of our QuantiDCE, which enables the trained metric to have obviously stronger correlations with human judgements than the other state-of-the-art metrics.


\begin{figure*}[t] 
	\centerline{\includegraphics[width=1\linewidth]{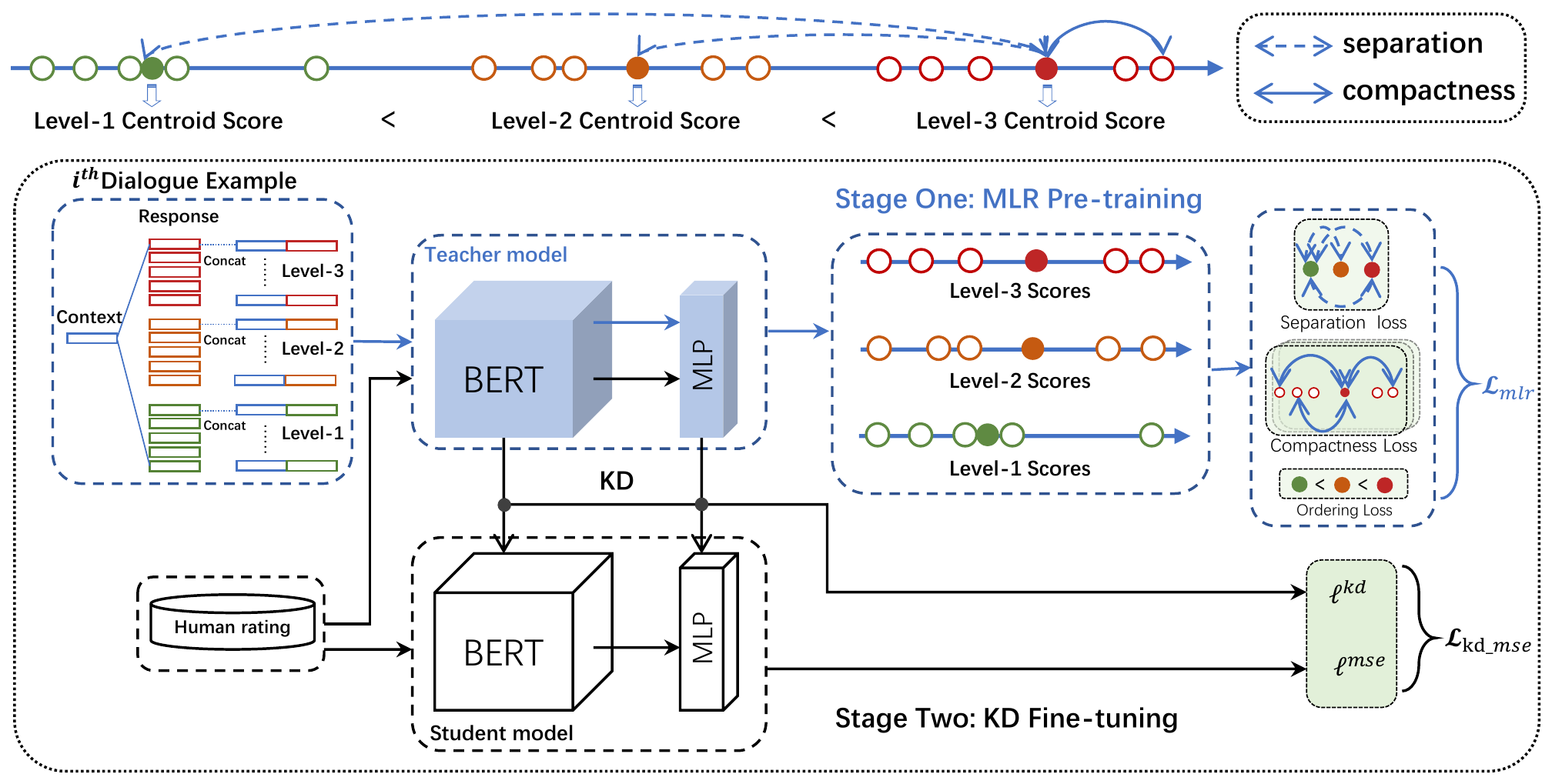}}
	\vspace{-3mm}
    \caption{The overall pipeline of our QuantiDCE, consisting of two training stages which are marked by the \textbf{\textcolor[RGB]{68,114,196}{blue}} and the \textbf{\textcolor{black}{black}} one-way arrows. Each input dialogue example contains one context with three-level candidate responses and five responses for each level, shown as \textbf{\textcolor[RGB]{190,38,38}{red}}, \textbf{\textcolor[RGB]{197,90,17}{orange}} and \textbf{\textcolor[RGB]{95,138,66}{green}} rectangles respectively. The solid circle represents the centroid score for each level of the $i^{th}$ dialogue. At MLR pre-training stage, the context-response pairs are encoded with BERT and transformed into the coherence scores through the MLP prediction network, and then MLR loss is applied to optimize the network. The dotted two-way arrows indicate that both ends should be separated, while the solid two-way arrows indicate that both ends should be compact. And at the KD fine-tuning stage, the student model is first initialized with the teacher model and optimized by KD-MSE loss.}
	\label{fig:arch}
	\vspace{-3mm}
\end{figure*}

\section{Related Work}
\label{sec:related_work}
    \paragraph{Automatic Coherence Evaluation.} 
    The widely used automatic metrics, such as BLEU~\citep{bleu}, METEOR~\citep{meteor} and ROUGE~\citep{rouge}, use statistical rules to measure the degree of lexical word-overlap between generated responses and reference responses. However, these metrics have been demonstrated to correlate poorly with human judgments due to the absence of semantic information~\citep{liu-etal-2016-evaluate,novikova-etal-2017-need}. Therefore, the subsequent metrics are considered to incorporate the semantic information. For instance, BERTScore~\citep{bert-score} turns to measure the soft semantic word-overlap rather than the hard lexical word-overlap like BLEU. Moreover, learnable metrics encoding the semantic information have been attracting interests recently, which are trained in a supervised manner with large-scale human-annotated data, such as ADEM~\citep{adem}, or trained in an unsupervised manner with automatically constructed data, such as RUBER~\citep{ruber} and BERT-RUBER~\citep{bert-ruber}. Besides, \citet{mesgar-etal-2020-dialogue} implicitly utilizes dialogue act labels by introducing an auxiliary task, which improves the performance of coherence evaluation. Furthermore, the recently proposed coherence metric, GRADE~\citep{grade}, introduces the graph information of dialogue topic transitions and achieves the current state-of-the-art results. Note that these learnable metrics are trained in a two-level training objective to separate the coherent dialogues from the incoherent ones, while our QuantiDCE models the task in a multi-level setting which is closer to the actual human rating.
    

    \paragraph{Knowledge Distillation.} Knowledge distillation (KD) is a method that transfers the knowledge from a large trained teacher model to a smaller student model by using the soft targets provided by the teacher~\citep{hinton2015}. In recent years, KD has been applied to many specific tasks~\citep{sun-etal-2020-distill,wei-etal-2019-online,kim-rush-2016-sequence,sourty-etal-2020-knowledge}. Unlike these previous works, we use KD to retain knowledge learned at the pre-training stage during fine-tuning and do not compress the model size of the student model.


\section{QuantiDCE Framework}
\label{sec:framework}

 
    In this section, we present QuantiDCE, a two-stage framework for dialogue coherence metric learning, consisting of Multi-Level Ranking (MLR) pre-training and Knowledge Distillation (KD) fine-tuning.
    As illustrated in Figure~\ref{fig:arch}, given a metric model $M$ (Section~\ref{subsec:model_architecture}), QuantiDCE enables $M$ to learn multi-level representations for context-response pairs with different levels of coherence degrees during the pre-training stage (Section~\ref{subsec:mlr_pretraining}), and further to learn the rating standards of humans with only a fraction of data during the fine-tuning stage (Section~\ref{subsec:kd_finetuning}). After these two training stages, the quantifiable gap between automatic metrics and humans can be obviously reduced.
    
    
    
    \subsection{Model Architecture}
    \label{subsec:model_architecture}
    In our QuantiDCE framework, the metric model $M$ is composed of: (1) an encoder network for encoding the input context-response pairs into features and (2) a predictor network for transforming the encoded features into coherence scores. Specifically, we adopt BERT~\citep{bert} as the encoder network and a multi-layer perceptron (MLP) as the predictor network.
    
    Given a context $\bm{c}=\{c_1, \cdots, c_m\}$ and a response $\bm{r}=\{r_1, \cdots, r_n\}$ where $c_i$ and $r_i$ are tokens of the context and the response respectively, the $\bm{c}$ and $\bm{r}$ are concatenated as $\{[{\rm CLS}], c_1, \cdots, c_m, [{\rm SEP}], r_1, \cdots, r_n, [{\rm SEP}]\}$, denoted as $[\bm{c};\bm{r}]$. Then the coherence score $\hat{s}$ of the response $\bm{r}$ w.r.t.\ the context $\bm{c}$ is predicted by:
    \begin{equation}
    \label{equation:forward}
        \hat{s} = MLP(BERT([\bm{c};\bm{r}])),
    \end{equation}
    where $MLP$ is a three-layer fully-connected network in which the activation functions of the three layers are two exponential linear
    units~\citep{elu} and a sigmoid function, respectively.
    
    \subsection{MLR Pre-Training}
    \label{subsec:mlr_pretraining}
    For learning the coarse judgement of coherence degrees without the direct supervision of score annotations, the model $M$ is first pretrained by minimizing a new multi-level ranking (MLR) loss on a large-scale dialogue dataset. Concretely, the MLR loss is composed of a separation loss, a compactness loss and an ordering loss.

    Formally, given a training dataset $\mathcal{D}_{pt} = \{(\bm{c}_i, \mathcal{R}_i)\}_{i=1}^{N_1}$ where $\bm{c}_i$ is a dialogue context and $\mathcal{R}_i=\{(\bm{r}_{i,1}^j, \cdots, \bm{r}_{i,K}^j)\}_{j=1}^{L}$ is a response set with $L$ coherence levels\footnote{The coherence level is in ascending order, i.e., the response in a higher level is more coherent than the lower one.} and $K$ responses for each level, the model $M$ is trained by minimizing the following MLR loss:
    \begin{equation}
    \label{equation:mlr_loss}
        \mathcal{L}_{mlr} = \frac{1}{N_1} \sum_{i=1}^{N_1} (\ell^{sep}_i + \ell^{com}_i + \ell^{ord}_i),
    \end{equation}
    where $\ell^{sep}_i$, $\ell^{com}_i$, and $\ell^{ord}_i$ refer to the separation loss, the compactness loss and the ordering loss of the $i^{th}$ example, respectively.

    \paragraph{The separation loss} aims to separate the features of context-response pairs with different coherence levels by separating the coherence scores of the different pairs\footnote{We also tried to directly restrict the features of different-level pairs to be separated, but the performance dropped compared with restricting the scores.}. Moreover, to efficiently compute the loss, we first compute the centroids of the context-response pairs belonging to the same coherence level for the $i^{th}$ dialogue example, i.e., $\bm{e}_i = \{e_i^j = \sum_{k=1}^{K} \hat{s}_{i,k}^j | j \in [1, L], e_i^j \in \mathbb{R}\}$ where $\hat{s}_{i,k}^j$ is the coherence score of the context-response pair ($\bm{c}_i$, $\bm{r}_{i,k}^j$), and the separation loss between the centroids is then computed as follows:
    \begin{equation}
    \label{equation:separation_loss}
        \ell^{sep}_i = \sum_{j=1}^{L-1} \sum_{l=j+1}^{L} max(0, w * \lambda - {\rm d}(e_i^j, e_i^l)),
    \end{equation}
    where ${\rm d}(\cdot)$ is the L1 distance, $\lambda$ is the lower bound for the distance between two centroids, and $w$ = $l-j$ is the distance weight used for amplifying the lower bound w.r.t. the coherence-level gap.

    \paragraph{The compactness loss} aims to compact the pairs within the same level, which served as a regularization role to avoid the occurrence of outliers for each coherence level. Specifically, the coherence score $\hat{s}_{i,k}^j$
    is forced to be closer to the corresponding centroid $e_i^j$ as follows:
    \begin{equation}
    \label{equation:compactness_loss}
        \ell^{com}_i = \sum_{j=1}^{L} \sum_{k=1}^{K} max(0, {\rm d}(e_i^j, \hat{s}_{i,k}^j) - \mu),
    \end{equation}
    where $\mu$ is the upper bound for the distance between the centroid of a certain coherence level and the score within this level.

    \paragraph{The ordering loss} is finally introduced to assure that the rank order of the predicted scores satisfies the pre-defined order of coherence degrees, i.e., $\hat{s}_{i,k}^j < \hat{s}_{i,k}^{j+1}$, $j\in[1, L-1]$, $k\in[1, K]$. It is critical since the separation loss only restricts the scores of the pairs from different coherence levels to be separated and this restriction is also satisfied when the scores of the highest level are lower than the scores of the lowest level. Similar to the separation loss, the ordering loss is also computed between each two centroids as follows: 
    \begin{equation}
    \label{equation:ordering_loss}
        \ell^{ord}_i = \sum_{j=1}^{L-1} \sum_{l=j+1}^{L} max(0, e_i^l - e_i^j).
    \end{equation}

    \subsection{KD Fine-Tuning}
    \label{subsec:kd_finetuning}
    The model $M$ pretrained by the MLR loss is further trained at the KD fine-tuning stage to directly learn the actual human rating standards with only a fraction of annotated data.

    Formally, given a training dataset $\mathcal{D}_{ft} =\{(\bm{c}_i, \bm{r}_i, s_i)\}_{i=1}^{N_2}$ where $\bm{c}_i$, $\bm{r}_i$ and $s_i$ are the dialogue context, the corresponding response and the human-annotated coherence score of $\bm{r}_i$ w.r.t. $\bm{c}_i$ respectively, the previous fine-tuning approach for the scoring task usually optimizes the model $M$ with an MSE loss between the predicted score $\hat{s}_i$ and the human score $s_i$:
    
    \begin{equation}
    \label{equation:mse_loss}
    \ell^{mse}_i = (s_i - \hat{s}_i)^2.
    \end{equation}

    However, by minimizing $\ell^{mse}_i$ for each example, the model $M$ will be easily over-fitting on the very few annotated data, and thus the model generalizability will be dramatically reduced. To overcome this issue, a novel knowledge distillation (KD) regularization is introduced for retaining the  knowledge learned at the MLR pre-training stage. Concretely, the pretrained model $M$ is treated as the teacher model that provides the soft targets for the student model $\hat{M}$ which is entirely copied from $M$. And we adopt the distillation objectives of TinyBERT~\citep{tinybert}, including the distillations of the embedding layer, the Transformer layers and the prediction layer. The KD loss is then formulated as:
    \begin{equation}
    \label{equation:kd_loss}
        \ell^{kd}_i = \sum_{t=0}^{T+1} |\!|O_i^{t} - \hat{O}_i^{t}|\!|_2^2 + \sum_{t=1}^{T} |\!|A_i^{t} - \hat{A}_i^{t}|\!|_2^2,
    \end{equation}
    where $|\!|\cdot|\!|_2^2$ indicates the squared L2 norm, T is the number of the Transformer layers, $O_i^{t}$ and $\hat{O}_i^{t}$ are the $t^{th}$ layer outputs of $M$ and $\hat{M}$ respectively, $A_i^{t}$ and $\hat{A}_i^{t}$ are the attention matrices of the $t^{th}$ transformer layer. Note that the layer 0 and the layer T+1 refer to the embedding layer and the prediction layer respectively.

    Overall, the loss function for KD fine-tuning, named as KD-MSE loss, is the weighted sum of $\ell^{mse}_i$ and $\ell^{kd}_i$ across the whole training dataset $\mathcal{D}_{ft}$:
    \begin{equation}
    \label{equation:kd_mse_loss}
        \mathcal{L}_{kd\_mse} = \frac{1}{N_2} \sum_{i=1}^{N_2} (\alpha * \ell^{mse}_i + \beta * \ell^{kd}_i),
    \end{equation}
    where $\alpha$ and $\beta$ are hyperparameters, and we empirically found that $\alpha$ = 1 and $\beta$ = 5 performs well.
    
    The overall training procedure is summarized in Algorithm~\ref{algorithm:1}.

    \begin{algorithm}[t]
        \caption{Training Procedure of QuantiDCE} 
        \hspace*{0.02in} {\bf Input:} training datasets $\mathcal{D}_{pt}$ and $\mathcal{D}_{ft}$, metric model $M$\\
        \hspace*{0.02in} {\bf Output:} student model $\hat{M}$
        \begin{algorithmic}[1]
        \State initialize $M$ with $\rm{BERT_{BASE}}$
        \ForAll{$(\bm{c}_i, \mathcal{R}_i) \in \mathcal{D}_{pt}$ }
                \State $\mathcal{S}_i = M(\bm{c}_i, \mathcal{R}_i)$
                \State compute the centroids $\bm{e}_i$ for $\mathcal{S}_i $
                \State compute $\ell^{sep}_i$ and $\ell^{ord}_i$ for $\bm{e}_i$
                \State compute $\ell^{com}_i$ between $\bm{e}_i$ and $\mathcal{S}_i$
            \State compute $\mathcal{L}_{mlr}$
            \State update $M$ to minimize $\mathcal{L}_{mlr}$
        \EndFor
        \State initialize $\hat{M}$ with $M$
        \ForAll{$(\bm{c}_i, \bm{r}_i, s_i) \in \mathcal{D}_{ft}$ }
            \State $O_i, A_i = M(\bm{c}_i, \bm{r}_i)$
            \State $\hat{s}_i, \hat{O}_i, \hat{A}_i = \hat{M}(\bm{c}_i, \bm{r}_i)$
            \State compute $\ell^{mse}_i$ between $s_i$ and $\hat{s}_i$
            \State compute $\ell^{kd}_i$ between $O_i, A_i$ and $\hat{O}_i, \hat{A}_i$
        \State compute $\mathcal{L}_{kd\_mse}$
        \State update $\hat{M}$ to minimize $\mathcal{L}_{kd\_mse}$
        \EndFor
        \State \Return student model $\hat{M}$
        \end{algorithmic}
        \label{algorithm:1}
    \end{algorithm}
    
\section{Experiments}
\label{sec:experiments}
    \subsection{Experimental Setup}
        \paragraph{Baseline Metrics.} We compare the metric model trained by our QuantiDCE with eight popular automatic dialogue metrics, including three lexical word-overlap metrics: BLEU~\citep{bleu}, ROUGE~\citep{rouge} and METEOR~\citep{meteor}, one semantic word-overlap metric, BERTScore~\citep{bert-score}, and four learnable metrics: ADEM~\citep{adem}, BERT-RUBER~\citep{bert-ruber}, BLEURT~\citep{bleurt} and GRADE~\citep{grade}.
        
        \paragraph{Evaluation.} Our QuantiDCE and the baselines are evaluated by computing the correlations between the model-predicted scores and the human-rated scores. Specifically, we adopt Pearson, Spearman and Kendall as the correlation measures and a large-scale human judgement benchmark~\citep{grade} to provide the human-rated scores.
        This benchmark contains 1,200 unique (context, response, human-rated score) triplets for metric evaluation where the contexts were randomly selected from the test set of three chit-chat datasets including DailyDialog~\citep{dailydialog}, ConvAI2~\citep{convai2} and EmpatheticDialogues~\citep{empathetic-dialogue}, and the responses were produced by both the retrieval-based dialogue models and the generation-based ones to assure response diversity.
        
        \paragraph{Training Datasets.}
        We use two datasets, DailyDialog++\footnote{\url{https://github.com/iitmnlp/Dialogue-Evaluation-with-BERT}} and DailyDialogEVAL\footnote{\url{https://github.com/li3cmz/GRADE}}, to support the pre-training and fine-tuning of QuantiDCE, respectively. The DailyDialog++ dataset~\citep{dailydialog++} contains over 11K conversations, which augments the original DailyDialog dataset with multiple responses of different quality levels including five golden reference responses, five adversarial irrelevant responses and five random selected responses for each context. Therefore, in this work, we set the number of coherence levels $L$ = 3 where the pairs containing the random responses, the adversarial responses and the reference responses respectively belong to the levels from 1 to 3. As to the fine-tuning data, we use the DailyDialog human judgement dataset, denoted as DailyDialogEVAL, which is a subset of the adopted evaluation benchmark~\citep{grade}, with 300 human rating data in total, and randomly split the data into training (90\%) and validation (10\%) sets.
        \paragraph{Implementation Details.} We use $\rm{BERT_{BASE}}$ to initialize the encoder network, which is in line with the current SOTA metric, GRADE. For the MLR pre-training, we pretrain our model for 5 epochs with batch size 3 and learning rate 2e-5 where the lower bound for the separation loss $\lambda$ = 0.3 and the upper bound for the compactness loss $\mu$ = 0.1. For the KD fine-tuning, we further finetune the pretrained model for 20 epochs with batch size 10 and learning rate 5e-6. For all the training, BERTAdam is used as the optimizer with $\beta_{1}=0.9$ and $\beta_{2}=0.999$. For the Transformer-layer distillation, we distill all the Transformer layers since the model architectures of the teacher and the student are exactly the same.

    \begin{table}[tbp]
    \centering
    \resizebox{.48\textwidth}{!}{
    \begin{tabular}{l|cccc}
    \hline
    Metric & Pearson & Spearman & Kendall & Average \\
    \hline
    \multicolumn{5}{c}{\textit{ConvAI2}}\\\hline
    BLEU & 0.003~\small{\textit{*}} & 0.128 & 0.088 & 0.073 \\

    ROUGE & 0.136 & 0.140 & 0.097 & 0.124 \\

    METEOR & 0.145 & 0.181 & 0.123 & 0.15 \\
    BERTScore & 0.225 & 0.225 & 0.154 & 0.201 \\
    \hline
    
    ADEM & 0.026~\small{\textit{*}} & 0.037~\small{\textit{*}} & 0.049~\small{\textit{*}} & 0.037 \\
    
    
    BERT-RUBER & 0.266 & 0.266 & 0.185 & 0.239 \\

    BLEURT & 0.152 & 0.149 & 0.103 & 0.135 \\

    GRADE & 0.496 & 0.503 & 0.356 & 0.452 \\
    \hline
    QuantiDCE & \textbf{0.554} & \textbf{0.554} & \textbf{0.395} & \textbf{0.501} \\
    \hline

    \multicolumn{5}{c}{\textit{EmpatheticDialogues}}\\\hline
    BLEU & -0.051~\small{\textit{*}} & 0.002~\small{\textit{*}} & 0.005~\small{\textit{*}} & -0.015 \\

    ROUGE & 0.029~\small{\textit{*}} & -0.013~\small{\textit{*}} & -0.010~\small{\textit{*}} & 0.002 \\

    METEOR & 0.118 & 0.055~\small{\textit{*}} & 0.04~\small{\textit{*}} & 0.071\\
    BERTScore & 0.046~\small{\textit{*}} & 0.033~\small{\textit{*}} & 0.021~\small{\textit{*}} & 0.033 \\
    \hline
    
    ADEM & 0.007~\small{\textit{*}} & 0.009~\small{\textit{*}} & 0.040~\small{\textit{*}} & 0.019 \\
    
    
    BERT-RUBER & -0.022~\small{\textit{*}} & -0.040~\small{\textit{*}} & -0.029~\small{\textit{*}} & -0.030 \\

    BLEURT & 0.203 & 0.192 & 0.13 & 0.175 \\

    GRADE & 0.350 & 0.344 & 0.243 & 0.312 \\
    \hline
    QuantiDCE & \textbf{0.412} & \textbf{0.393} & \textbf{0.274} & \textbf{0.360} \\
    \hline
    \end{tabular}
    }
    \vspace{-2mm}
    \caption{Correlations between automatic evaluation metrics and human judgements on two datasets (ConvAI2 and EmpatheticDialogues). The star~\small{\textit{*}} indicates results with p-value $>$ 0.05, which are not statistically significant.} 
    \label{table:metric_performance}
    \vspace{-3mm}
    \end{table}
    
    \begin{table}[t]
    \centering
    \resizebox{.48\textwidth}{!}{
    \begin{tabular}{l|cccc}
    \hline
    Loss & Pearson & Spearman & Kendall & Average \\
    \hline
    \multicolumn{5}{c}{\textit{ConvAI2}}\\\hline

    BCE & 0.505 & 0.505 & 0.361 & 0.457 \\
    Ranking & 0.507 & 0.504 & 0.360 & 0.457 \\
    \hline
    SupCon & 0.495 & 0.523 & 0.367 & 0.462 \\
    FAT & 0.516 & 0.521 & 0.371 & 0.469 \\
    Vanilla MLR & 0.522 & 0.536 & 0.379 & 0.479 \\
    \hline
    MLR (ours) & \textbf{0.554} & \textbf{0.554} & \textbf{0.395} & \textbf{0.501} \\
    \hline

    \multicolumn{5}{c}{\textit{EmpatheticDialogues}}\\\hline

    BCE & 0.354 & 0.353 & 0.243 & 0.317 \\
    Ranking & 0.399 & 0.389 & 0.272 & 0.353 \\
    \hline
    SupCon & 0.332 & 0.315 & 0.22 & 0.289 \\
    FAT & 0.381 & 0.358 & 0.245 & 0.328 \\
    Vanilla MLR & 0.403 & 0.387 & 0.267 & 0.352 \\
    \hline
    MLR (ours) & \textbf{0.412} & \textbf{0.393} & \textbf{0.274} & \textbf{0.360} \\
    \hline

    \end{tabular}
    }
    \vspace{-2mm}
    \caption{Correlations between human judgements and the metric models trained with different losses during pre-training and the same KD-MSE loss during fine-tuning. Ranking represents the margin ranking loss.} 
    \label{table:pre-training_objective}
    \vspace{-3mm}
    \end{table}

    \subsection{Experimental Results}
    \paragraph{Metric Performance.} The correlation results of QuantiDCE and the other baseline metrics on the large-scale human judgement benchmark are presented in Table~\ref{table:metric_performance}, including the ConvAI2 and the EmpatheticDialogues datasets.\footnote{The DailyDialogEVAL dataset was not used for evaluation since we used it for fine-tuning.}
    For a fair comparison, the learnable baseline metrics, ADEM, BERT-RUBER and GRADE, are trained on the training dataset we adopted, i.e., DailyDialog++.\footnote{BLEURT was not trained on DailyDialog++ since this dataset is not suitable for the BLEURT pre-training strategy. Instead, we trained BLEURT with the fine-tuning data we used. The training details of these baseline metrics are provided in Appendix \ref{appendix:details_baseline}.}
    Generally, QuantiDCE achieves an absolute averaged correlation improvement by around 5\% points over the current SOTA, GRADE. Besides, all the results of QuantiDCE are statistically significant with p-value \textless 0.01.

    \begin{figure*}[t] 
    	\centerline{\includegraphics[width=1\linewidth]{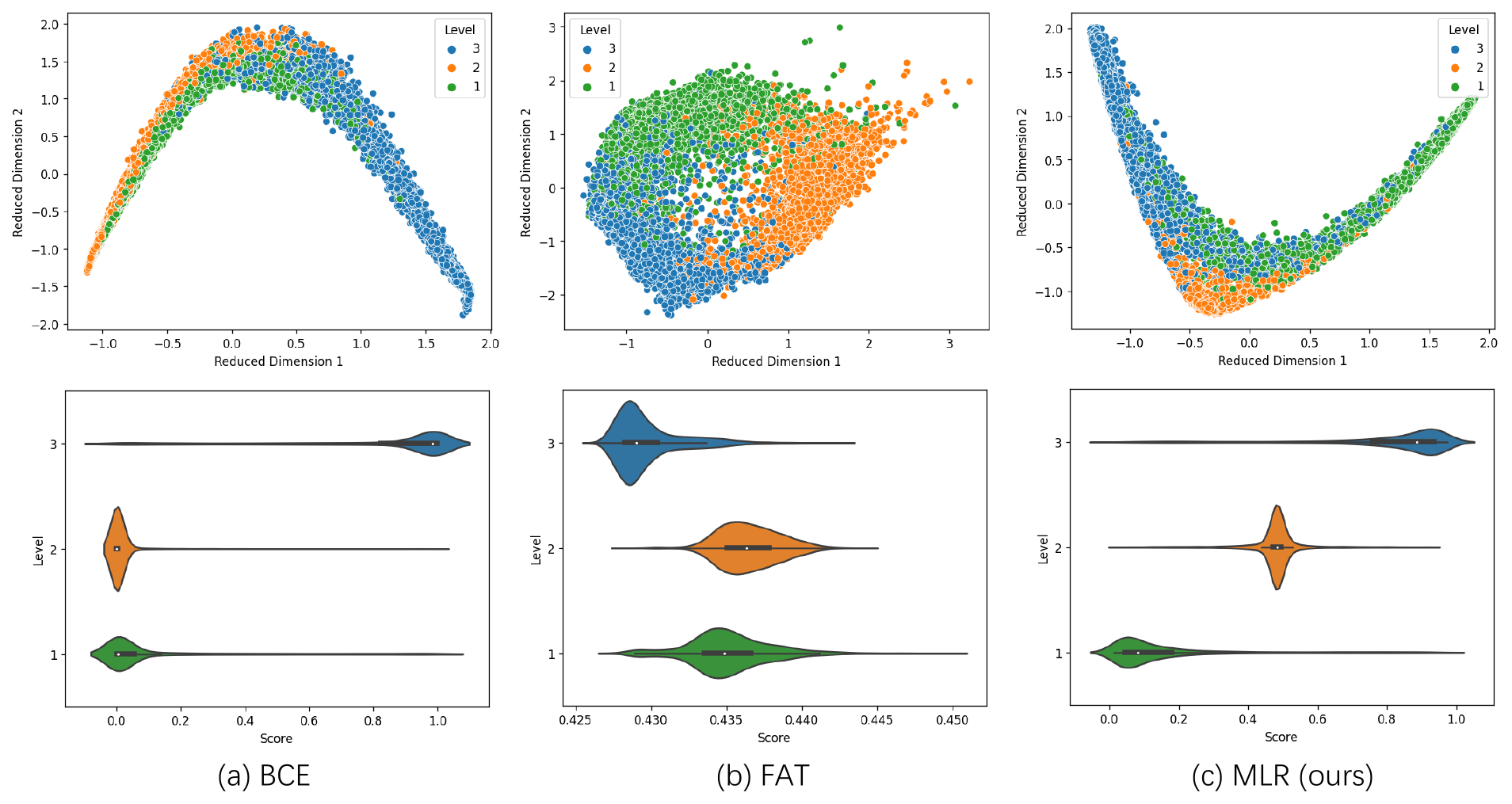}}
    	\vspace{-3mm}
    	\caption{Visualizations of features (the scatter plots in the upper row) and scores (the violin plots in the lower row) on the dailydialog++ dataset. The features and scores in each of the three columns are obtained from the metric model $M$ only pretrained with the BCE loss, the FAT loss and our MLR loss, respectively.}
    	\label{fig:visualization}
    	\vspace{-3mm}
    \end{figure*}

    \paragraph{Pre-Training Objective.} To verify the superiority of our pre-training objective, namely the MLR loss, we investigated the performance of several existing loss functions for pre-training compared with ours. Specifically, two categories of loss functions used for metric training are adopted, including (a) the two-level setting and (b) the multi-level setting. The binary cross entropy (BCE) loss and the margin ranking loss are adopted for the two-level setting, while another three loss functions are adopted for the multi-level setting, including the supervised contrastive (SupCon) loss~\citep{khosla2020supervised}, the fast-approximated triplet (FAT) loss~\citep{yuan2019fat} and the vanilla MLR loss~\citep{wordNotBi} \footnote{The details of these pre-training loss fucntions are provided in Appendix \ref{appendix:details_pretraining_loss}.}. As shown in Table~\ref{table:pre-training_objective}, the performance of our MLR loss is the best among all the pre-training objectives. And we also found that the multi-level setting losses perform better than the two-level ones, especially on the ConvAI2 dataset. Moreover, in order to more intuitively analyze the performances of these pre-training objectives, we also visualize the encoded features and the predicted scores of the model $M$ after being pretrained by the above loss functions on the DailyDialog++ dataset without fine-tuning.\footnote{The visualization results of the ranking loss, SupCon loss and Vanilla MLR loss are provided in Appendix \ref{appendix:visualize_pretraining_loss}.} As shown in Figure~\ref{fig:visualization}, (a) the BCE loss cannot separate the level-1 scores from the level-2 ones and the corresponding features are also mixed; (b) the FAT loss, on the other hand, separates the features of different levels well, but does not consider the relative gaps where the distances between the level-1 and level-3 features are not larger than those between level-1 and level-2; (c) in contrast, our MLR loss separates both the features and the scores well and also considers the relative gaps between different levels.

    \begin{table}[t]
    \centering
    \resizebox{.48\textwidth}{!}{
    \begin{tabular}{l|cccc}
    \hline
    Loss & Pearson & Spearman & Kendall & Average \\
    \hline

    \multicolumn{5}{c}{\textit{ConvAI2} ({\rm best epoch})}\\\hline
    MSE & 0.272 & 0.369 & 0.255 & 0.299 \\
    MSE (fix encoder) & 0.477 & 0.477 & 0.337 & 0.430 \\
    \hline
    KD-MSE (ours) & \textbf{0.554} & \textbf{0.554} & \textbf{0.395} & \textbf{0.501} \\
    \hline

    \multicolumn{5}{c}{\textit{EmpatheticDialogues} ({\rm best epoch})}\\\hline
    MSE & 0.278 & 0.276 & 0.187 & 0.247 \\
    MSE (fix encoder) & 0.384 & 0.367 & 0.253 & 0.335 \\
    \hline
    KD-MSE (ours) & \textbf{0.412} & \textbf{0.393} & \textbf{0.274} & \textbf{0.360} \\
    \hline

    \multicolumn{5}{c}{\textit{DailyDialogEVAL} ({\rm last epoch})}\\\hline
    MSE & 0.934 & 0.945 & 0.867 & 0.915 \\
    MSE (fix encoder) & 0.379 & 0.402 & 0.281 & 0.354 \\
    \hline
    KD-MSE (ours) & 0.804 & 0.832 & 0.678 & 0.771 \\
    \hline

    \end{tabular}
    }
    \vspace{-2mm}
    \caption{Correlations between human judgements and the metric model $M$ further trained with different fine-tuning losses after MLR pre-training.} 
    \label{table:fine_tune_strategy}
    \vspace{-3mm}
    \end{table}

    \begin{figure}[t] 
    	\centerline{\includegraphics[width=0.9\linewidth]{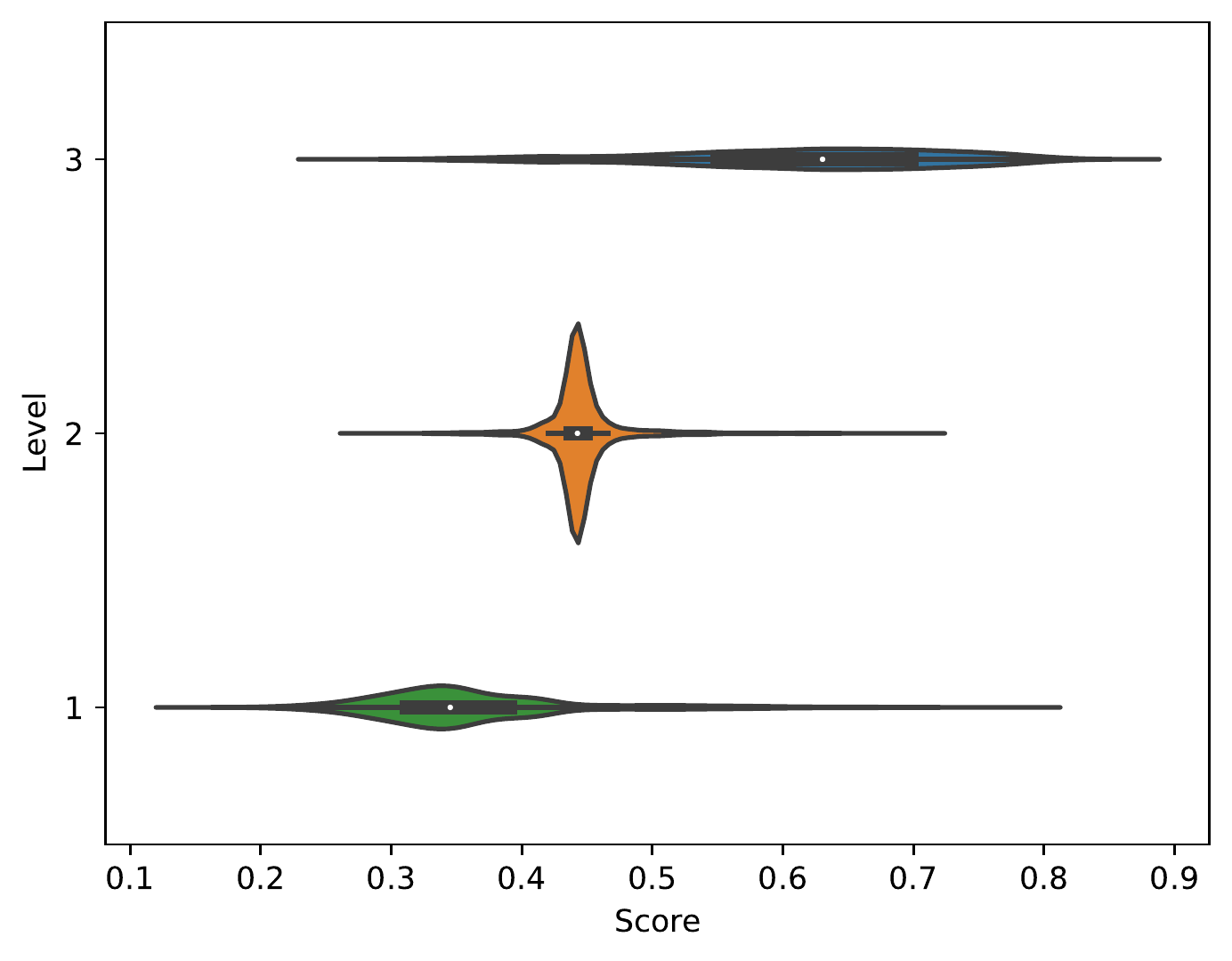}}
    	\vspace{-3mm}
    	\caption{Score visualization on the dailydialog++ dataset where the scores are predicted by our QuantiDCE after KD fine-tuning.}
    	\label{fig:score_visualization_after_fine_tuning}
    	\vspace{-3mm}
    \end{figure}

    \paragraph{Fine-Tuning Objective.} Furthermore, we also verified the effectiveness of our KD-MSE loss during fine-tuning by comparing with other fine-tuning losses, including the pure MSE loss without KD regularization as shown in Equation~\ref{equation:mse_loss} and the same MSE loss except for freezing the encoder network and only finetuning the predictor network i.e. the MLP, denoted as MSE (fix encoder). As the results shown in Table~\ref{table:fine_tune_strategy}, compared with the other two losses, the model finetuned by our KD-MSE loss has the highest correlation results on both ConvAI2 and EmpatheticDialogues. Moreover, by comparing the results of MSE and KD-MSE, we can find that introducing KD regularization leads to obvious averaged correlation improvements by 20.2\% points on ConvAI2 and 11.3\% points on EmpatheticDialogues, which verifies the effectiveness of the KD loss. Besides, we also reported the last-epoch correlation results on the training dataset, DailyDialogEVAL. And the results of MSE and MSE (fix encoder) indicate the phenomena of over-fitting and under-fitting into DailyDialogEVAL respectively, which explain the reasons of their low performance on the two evaluation datasets. In contrast, our KD-MSE loss enables the model to learn the actual human rating standards from the scarce annotated data and avoid overfitting it simultaneously. Finally, in Figure~\ref{fig:score_visualization_after_fine_tuning}, we present the visualization of the scores predicted by our QuantiDCE after KD fine-tuning. Compared with the score distributions before fine-tuning in Figure~\ref{fig:visualization}(c), the finetuned score distributions of the level-1 and level-3 are wider and partly overlap with the level-2 distribution. It is predictable as the judgements of coherence are always subjective and humans tend to give vague and middle scores instead of extremely high or low scores.

    \begin{table}[tbp]
    \centering
    \resizebox{0.48\textwidth}{!}{
    \begin{tabular}{l|cccc}
    \hline
    Metric & Pearson & Spearman & Kendall & Average \\
    \hline
    QuantiDCE & \textbf{0.554} & \textbf{0.554} & \textbf{0.395} & \textbf{0.501} \\
    \hline
    w/o MLR pre-training & 0.373 & 0.357 & 0.246 & 0.325 \\
    \qquad w/o separation loss & 0.388 & 0.416 & 0.289 & 0.364 \\
    \qquad w/o compactness loss & 0.526 & 0.550 & 0.390 & 0.489 \\
    \qquad w/o ordering loss & -0.494 & -0.522 & -0.371 & -0.462 \\
    w/o KD fine-tuning & 0.531 & 0.540 & 0.381 & 0.484 \\
    \hline
    \end{tabular}
    }
    \vspace{-2mm}
    \caption{Ablation studies on the ConvAI2 dataset by removing one of the component in QuantiDCE, including the MLR loss (w/o MLR pre-training), the KD+MSE loss (w/o KD fine-tuning), and three secondary losses of the MLR loss.} 
    \label{table:ablation}
    \end{table}

    \begin{figure}[t] 
    	\centerline{\includegraphics[width=0.85\linewidth]{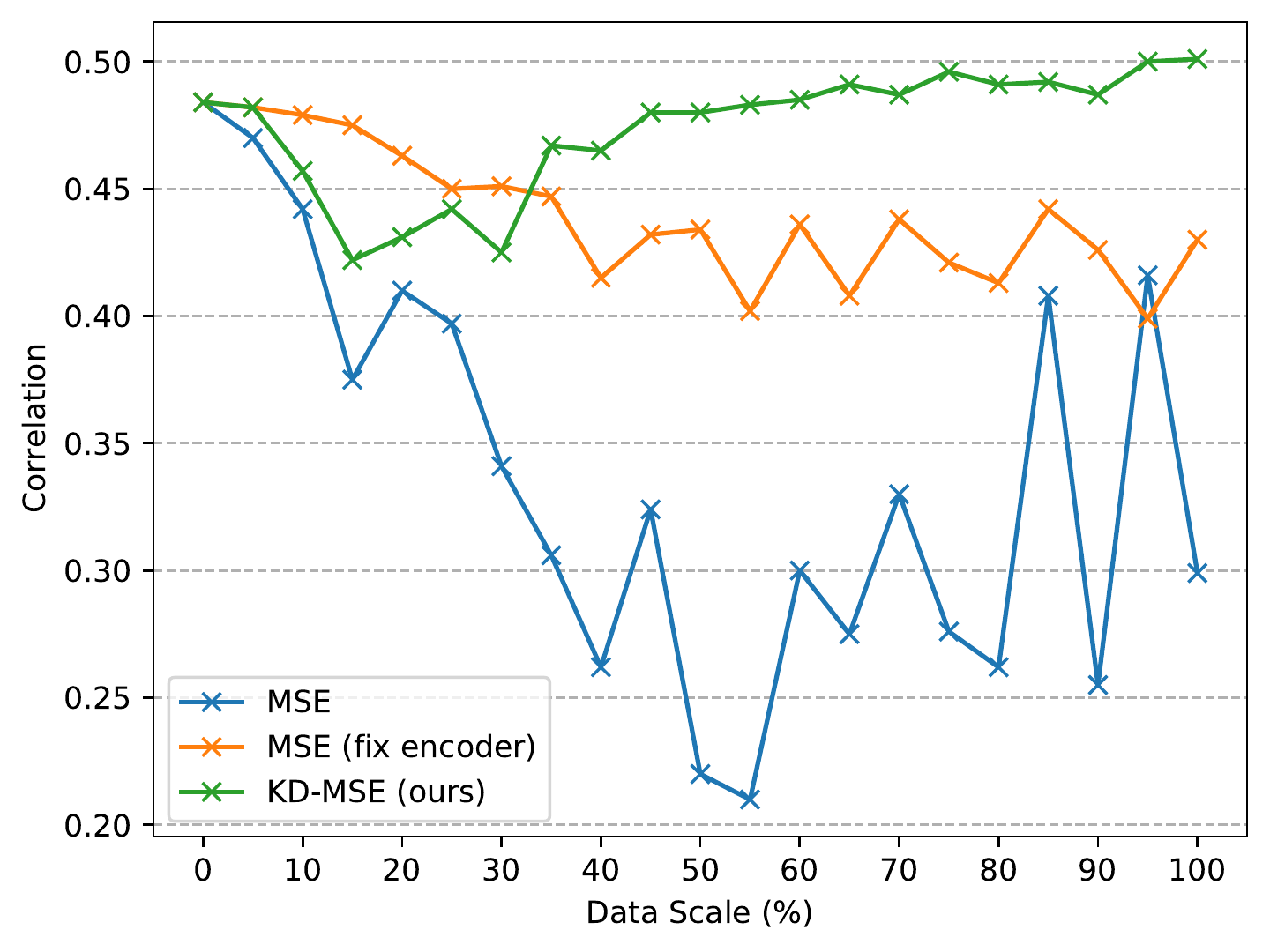}}
    	\vspace{-3mm}
    	\caption{The performance trends when changing the number of annotated data used for different fine-tuning objectives. Each point in the line chart indicates the averaged correlation of Pearson, Spearman and Kendall.}
    	\label{fig:number_of_data_for_fine-tuning}
    	\vspace{-3mm}
    \end{figure}

    \subsection{Ablation Studies}



        \paragraph{Component Analysis.} To verify the contributions of the core components in our QuantiDCE, we further conducted ablation studies on the ConvAI2 dataset. As shown in Table~\ref{table:ablation}, both the MLR pre-training and KD fine-tuning contribute to the better performance of QuantiDCE. Besides, we also conducted ablations by removing one of the secondary loss during MLR pre-training, including the separation loss, the compactness loss and the ordering loss. The results show that the performance benefits from all these losses in which the separation loss and the ordering loss are crucial for training a metric with strong and positive human correlations.

        \paragraph{Number of Data for Fine-Tuning.} Moreover, we also investigated how the scale of data for fine-tuning effects the model performance by increasing the number of fine-tuning data 5\% each time from zero. The trend of the model performance is presented in Figure~\ref{fig:number_of_data_for_fine-tuning}. We observed that minimizing our KD-MSE loss made the correlation results have a gradually increasing trend after an initial decrease.\footnote{The initial decrease probably attributes to the randomness of data sampling where the smaller the sampling ratio is, the higher the probability that noisy samples dominate the sampled data will be. And overfitting into the noisy samples leads to the performance decrease.} More specifically, the result achieved the standard before fine-tuning at around the 70\% data scale and continued increasing until 100\% with a final improvement by around 2\% points. For comparison, the performance trends of MSE and MSE (fix encoder) are also provided.
        And the results present overall decreasing trends of the model performance, which indicates that the model trained by MSE or MSE (fix encoder) cannot benefit from the increasing of data scale,
        due to the severe over-fitting or under-fitting.
        Therefore, to effectively utilize the limited data, it is important to enable the update of the entire network and add some constraints to avoid over-fitting, such as our proposed KD regularization.
        
        \begin{table}[t]
            \centering
            \resizebox{0.48\textwidth}{!}{
            \begin{tabular}{l}
            \toprule[1.5pt]
            \textbf{U1}: I need to book a plane ticket to London. \\
            \textbf{U2}: Round-trip or one-way? \\
            \textbf{R}:\quad Round trip or one way trip? \\
            Coherence Score (\textbf{\textcolor{red}{Human}} / \textbf{\textcolor[RGB]{34,139,34}{QuantiDCE}} / \textbf{\textcolor{blue}{GRADE}}) : \textcolor{red}{2.10} / \textcolor[RGB]{34,139,34}{2.85} / \textcolor{blue}{4.52} \\
            \hline
            \textbf{U1}: Yum. You will find me in the kitchen and if not i am fishing. \\
            \textbf{U2}: Wow that's pretty cool what else you do for fun? \\
            \textbf{R}:\quad Probably fish it is great. \\
            Coherence Score (\textbf{\textcolor{red}{Human}} / \textbf{\textcolor[RGB]{34,139,34}{QuantiDCE}} / \textbf{\textcolor{blue}{GRADE}}) : \textcolor{red}{2.50} / \textcolor[RGB]{34,139,34}{3.94} / \textcolor{blue}{4.27} \\
            \bottomrule[1.5pt]
            \end{tabular}
            }
            \vspace{-2mm}
            \caption{Two representative examples to show the strength and weakness of our QuantiDCE where U1 and U2 are two utterances of the context and R is the corresponding response.} 
            \label{table:case_study}
            \vspace{-3mm}
        \end{table}
        
        \subsection{Case Study}
        To illustrate the performance of QuantiDCE, two representative examples are shown in Table~\ref{table:case_study} . The first example shows the strength of QuantiDCE where the coherence score given by ours is closer to the human rating score compared with the extremely high score given by GRADE. However, in the second example, both our QuantiDCE and GRADE deviate from the human score, possibly because the number of coherence levels we adopted in this work ($L$ = 3) is insufficient as humans usually consider more levels of dialogue coherence.

    

\section{Conclusion}
\label{sec:conclusion}
In this paper, we propose QuantiDCE, a novel training framework aiming to bridge the gap between the training objective and the actual human rating and train a quantifiable dialogue coherence metric. In general, QuantiDCE includes two training stages, MLR pre-training for learning the coarse human judgements of dialogue coherence degrees, and KD fine-tuning for learning the actual human rating standards. Experimental results show that the metric trained by QuantiDCE presents strong correlations with human judgements. For future work, it is interesting to investigate a more efficient way to obtain multi-level data and extend the multi-level setting into the general evaluation for natural language generation.

\section*{Acknowledgments}
We thank all anonymous reviewers for their constructive comments. This work was supported in part by National Key R\&D Program of China under Grant No. 2020AAA0109700, National Natural Science Foundation of China (NSFC) under Grant No.U19A2073 and No.61976233, Guangdong Province Basic and Applied Basic Research (Regional Joint Fund-Key) Grant No.2019B1515120039,  Shenzhen Fundamental Research Program (Project No. RCYX20200714114642083, No. JCYJ20190807154211365), Zhijiang Lab’s Open Fund (No. 2020AA3AB14) and CSIG Young Fellow Support Fund.

\bibliographystyle{acl_natbib}
\bibliography{acl2021}

\appendix

\section{Training Details of the learnable baseline metrics}
\label{appendix:details_baseline}
     a) Following \citet{dailydialog++}, we trained ADEM by first initializing it with the official checkpoint and further finetuning on DailyDialog++ with a target of 5 for level-3 pairs and 1 for level-1 pairs; b) BERT-RUBER and GRADE were both trained on DailyDialog++ where level-3 pairs as positive samples and both level-1 and level-2 pairs as negative samples, except that the former use cross-entropy loss while the latter use ranking loss; c) BLEURT was initialized with the official recommended checkpoint BLEURT-Base and finetuned on DailyDilaogEVAL by following the office guidelines\footnote{\url{https://github.com/google-research/bleurt}}.

\section{Details of the Pre-Training Losses}
\label{appendix:details_pretraining_loss}

\paragraph{BCE Loss.}
    The binary cross entropy (BCE) loss is adopted for the experiments of the two-level setting, where both the adversarial irrelevant responses and random selected responses of the dailydialog++ dataset~\citep{dailydialog++} are treated as negative samples and labeled as 0, while the golden reference responses are treated as positive samples and labeled as 1.
    
\paragraph{Margin Ranking Loss.}
    Similarly, the margin ranking loss simplifies the evaluation task as a two-level setting and maximizes the differences between the positive coherent dialogues and the negative incoherent ones. As the name suggests, the focus of the margin ranking loss is ranking, which aims at ranking the scores of positive coherent dialogues ahead of the negative incoherent ones. 
    
\paragraph{SupCon Loss.}
    The supervised contrastive (SupCon) loss~\citep{khosla2020supervised}, which pulls the positive anchors closer and pushes the negatives farther away in representation space, can be adopted for the multi-level setting. Here, for our multi-level setting, we consider the dialogues of level-1, level-2, and level-3 as positive anchors successively, and the remaining two levels as corresponding negatives.
    
\paragraph{FAT Loss.}
    The fast-approximated triplet (FAT) loss~\citep{yuan2019fat} replaces the traditional point-to-point distances of the triplet loss with point-to-cluster distances, through an upper bound relaxation of the triplet form, which is first applied for the classification task and obviously reduces the computation cost. To use FAT loss in our evaluation task, we consider the different coherence levels as different classes and perform the FAT loss to separate the context-response pairs with different coherence levels.
    
\paragraph{Vanilla MLR Loss.}
    The vanilla MLR loss~\citep{wordNotBi} is the extension of the margin ranking loss to a multi-level version by repeatedly applying the original margin ranking loss between different levels, which can be directly applied to our evaluation task.

\section{Visualizations of the Pre-Training Losses}
\label{appendix:visualize_pretraining_loss}
    We have already compared the visualization results of the BCE loss and the FAT loss. For a supplement, here we mainly introduce the visualizations of the margin ranking loss, the SupCon loss and the vanilla MLR loss in detail.
    
\begin{figure*}[t] 
	\centerline{\includegraphics[width=1\linewidth]{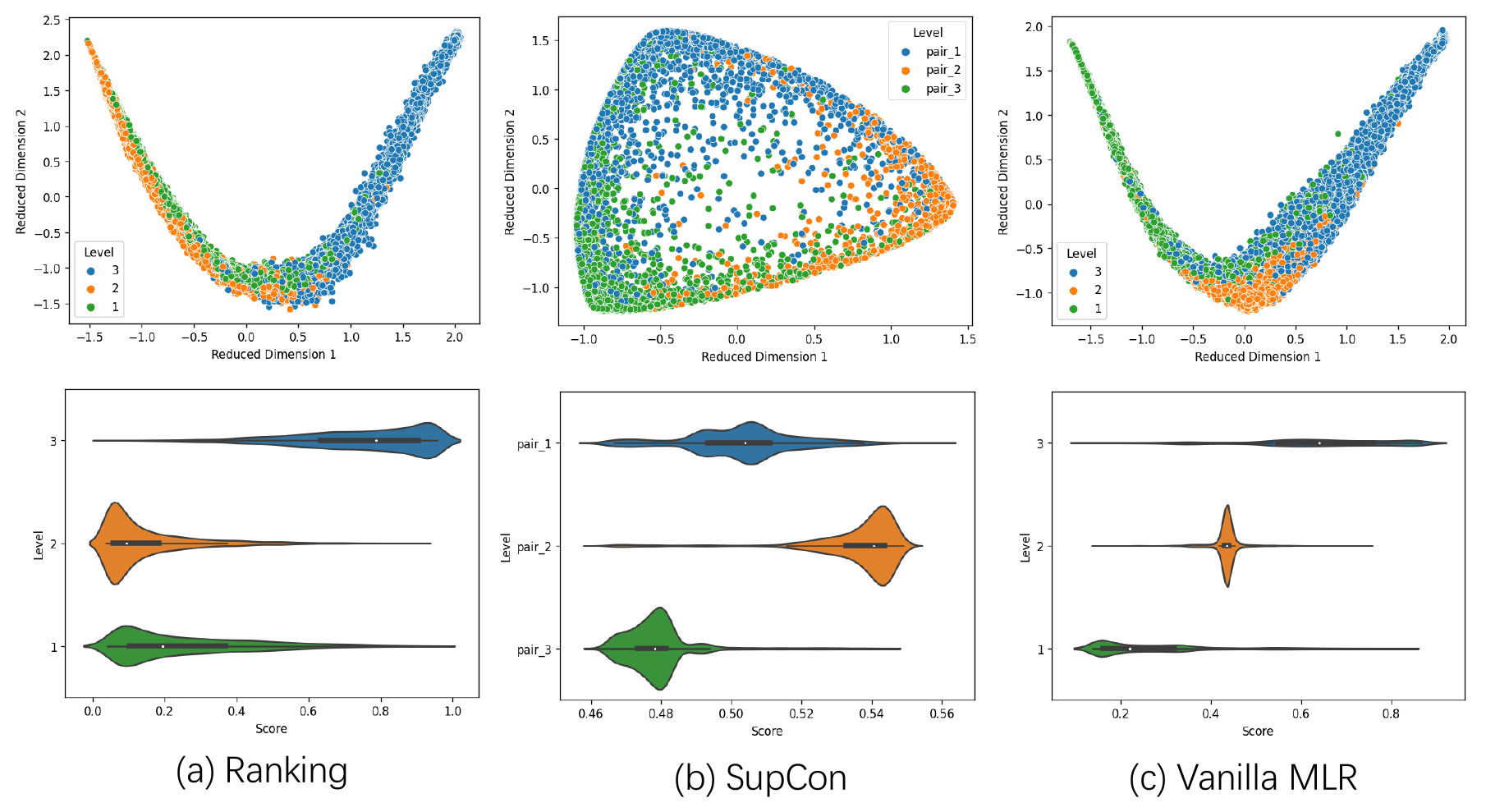}}
	\vspace{-3mm}
	\caption{Visualizations of features (the scatter plots in the upper row) and scores (the violin plots in the lower row) on the dailydialog++ dataset. The features and scores in each of the three columns are obtained from the metric model $M$ only pretrained with the margin ranking loss, the SupCon loss and the vanilla MLR loss, respectively.}
	\label{fig:visualization_supp}
	\vspace{-3mm}
\end{figure*}

    As we can see in Figure \ref{fig:visualization_supp}, (a) the margin ranking loss cannot separate the level-1 scores from the level-2 ones and the corresponding features are also mixed, which is similar to the BCE loss; (b) the SupCon loss, on the other hand, can distinguish the features and scores of the three levels to some extent, and the scores of different levels are also separated but do not follow the real rank order, i.e., level-1 $<$ level-2 $<$ level-3; (c) the final vanilla MLR loss can separate the context-response pairs with different coherence level in feature space and the predicted scores also follow the actual rank order. However, its score distributions are not compact enough for the level-1 and level-3.
   


\end{document}